\pdfoutput=1
\documentclass[letterpaper, 10 pt,conference]{ieeeconf}  

\IEEEoverridecommandlockouts                              

\title{\LARGE \bf
A Novel Gaussian Process Based Ground Segmentation
Algorithm with Local-Smoothness Estimation}
\author{Pouria Mehrabi and Hamid D. Taghirad,~\emph{Senior Member,~IEEE}.
\thanks{Authors are with the Advanced
Robotics and Automated Systems (ARAS),
Industrial Control Center of Excellence, Faculty of Electrical
Engineering, K. N. Toosi University of Technology, Tehran, Iran
(e-mail: pmehrabi@email.kntu.ac.ir and taghirad@kntu.ac.ir)}}

\usepackage{multicol}
\usepackage{color}
\usepackage{amsmath}
\usepackage{amssymb}
\usepackage{algorithm}
\usepackage[noend]{algpseudocode}
\usepackage[english]{babel}
\usepackage{adjustbox}
\usepackage{graphicx}
\usepackage{setspace}
\graphicspath{ {Images/} }
\usepackage{tikz}
\usetikzlibrary{fit,positioning}
\usetikzlibrary{automata}
\usetikzlibrary{arrows,intersections}
\usepackage{environ}
\usepackage{tkz-euclide}
\makeatletter
\newsavebox{\measure@tikzpicture}
\NewEnviron{scaletikzpicturetowidth}[1]{%
  \def\tikz@width{#1}%
  \begin{lrbox}{\measure@tikzpicture}%
  \BODY
  \end{lrbox}%
  \pgfmathparse{#1/\wd\measure@tikzpicture}%
  \BODY
}
\makeatother
 \usepackage{stackengine}
\stackMath

\usepackage{amsfonts}
\usepackage{booktabs}
\usepackage{siunitx}
\usepackage{tabularx}
\usepackage[margin=0.792in]{geometry}
\begin{document}
\maketitle \thispagestyle{empty} \pagestyle{empty}
\begin{abstract}
Autonomous Land Vehicles (ALV) shall efficiently recognize the
ground in unknown environments. A novel $\mathcal{GP}$-based method is proposed for the ground segmentation task in rough driving scenarios. The non-stationary covariance function proposed by \cite{Paciorek2003} is utilized as the kernel for the $\mathcal{GP}$. The ground surface behavior is assumed to only demonstrate local-smoothness. Thus, point estimates of the kernel's length-scales are obtained. Thus, two Gaussian processes are introduced to separately model the
observation and local characteristics of the data. While, the \textit{observation process} is used to model the ground, the \textit{latent process} is put on length-scale values to estimate point values of length-scales at each input location. Input locations for this latent process are chosen in a physically-motivated procedure to represent an intuition about
ground condition. Furthermore, an intuitive guess of length-scale value is represented by assuming the existence of hypothetical surfaces in the environment that every bunch of data points may be assumed to be resulted from measurements from this surfaces. Bayesian inference is implemented
using \textit{maximum a Posteriori} criterion. The log-marginal likelihood
function is assumed to be a multi-task objective function, to
represent a whole-frame unbiased view of the ground at each frame.
Simulation results shows the effectiveness of the proposed method
even in an uneven, rough scene which outperforms similar Gaussian process
based ground segmentation methods. While adjacent segments do not
have similar ground structure in an uneven scene, the proposed
method gives an efficient ground estimation based on a whole-frame viewpoint
instead of just estimating segment-wise probable ground surfaces.
\end{abstract}

\section{Introduction}
In \cite{Mehrabi2021} is shown that the efficient ground segmentation task for the sloped terrains needs to be addressed due to consideration of physical-motivated qualities of the data. For the rough scenes, this assumption should be developed more.
The technology trends shows more interest in autonomous land
vehicles (ALV) with the growth of research interest into the
subject. ALV's are able to provide many opportunities from
empowering the ability of remote exploration and navigation in an
unknown environment to establishing driver-less cars that are able
to navigate autonomously in urban areas being more safe by
compensating human driving faults. In order to establish a
driver-less car capable of performing autonomously in urban areas,
developed methods shall be reliable and real-time implementable.
Ground segmentation represents itself as a vital component of any
algorithm pursuing further tasks in an unknown environment. A
reliable ground segmentation procedure shall be applicable in
environments with both flat and sloped terrains, while being realistic
and real-time implementable, since often this task is only a
prerequisite for other time consuming algorithms and it is an important basic part of ALV's perception of it's surroundings. Gaussian process
regressions are useful tools for implementation of Bayesian
inference which relies on correlation models of inputs and
observation data \cite{C.E.Rasmussen2006}. They provide fast and
fully probabilistic framework for non-linear regression problems.
Although light detection and ranging (LIDAR) sensors are commonly
used in ALVs, data resulted from these sensors, does not inherit
smoothness, and therefore, stationary covariance functions may not
be used to implement Gaussian regression tasks on these data. Different methods for segmentation have been proposed in
the literature \cite{Zermas2017, Shin2017, Korchev2016, Asvadi2016,
Chen2014, Himmelsbach2010, Moosmann2009}.

In \cite{Zermas2017}, ground surface is obtained in an iterative
routine, using deterministically assigned seed points. In
\cite{Shin2017}, the ground segmentation step is put aside to
establish a faster segmentation based on Gaussian process
regression. A 2D occupancy grid is used to determine surrounding
ground heights, and furthermore, a set of non-ground candidate
points are generated. Reference \cite{Korchev2016} handles real-time
segmentation problem by differentiating the minimum and maximum
height map in both rectangular and a polar grid map. In
\cite{Asvadi2016} a geometric ground estimation is obtained by a
piece-wise plane fitting method capable of estimating arbitrary
ground surfaces. In \cite{Chen2014} a Gaussian process based
methodology is used to perform ground estimation by segmenting the
data with a fast segmentation method firstly introduced by
\cite{Himmelsbach2010}. The non-stationary covariance function from
\cite{Paciorek2003} is used to model the ground observations while
no specific physical motivated method is given for choosing
length-scales. Paper \cite{Moosmann2009} proposes a fast
segmentation method based on local convexity criterion in non-flat
urban environments.

These methods are either estimating ground piece-wise 
and with local viewpoint or by labeling all the individual points with some predefined criterion. Except \cite{Chen2014}
non of the methods above, gives a continuous model for predicted ground. Furthermore none of them gives an
exact, physical motivated routine to extract local characteristics
of non-smooth data, while efficient ground segmentation have to be
done considering physical realities of the data including non-smoothness of the LIDAR data and ground condition in every data frame. 

Due to the ability of Gaussian processes ($\mathcal{GP}$) to model correlations
between data points, there is a growth of interest seen in the
literature to use them with 3D point clouds. Furthermore,
$\mathcal{GP}$s are capable of estimating functional relationships
by considering correlations between observations and data points,
even when no model is available and the function is prone to huge
changes. The correlation is introduced to $\mathcal{GP}$s with
covariance kernels. Covariance functions are key concepts in
Gaussian process regressions as they define how data points relate
to each other. Specification of covariance structures is critical
specially in non-parametric regression tasks \cite{Stein2005}.

LIDAR data is consisted of three-dimensional range data which is
collected by a rotating sensor, strapped down to a moving car. This
moving  sensor obviously causes non-smoothness in it's measured data,
which may not be taken into account using common stationary
covariance functions. Although a Gaussian process based method for
ground segmentation with non-stationary covariance functions is
proposed in the literature to take input-dependent smoothness into
account in \cite{Chen2014}, adjusting covariance kernels to
accommodate with physical reality of the ground segmentation problem
needs further investigation. Length-scales may be defined as the
extent of the area that data points can effect on each other
\cite{Plagemann2008}. Length-scale values plays a significant role
in the quality of the interpretation that covariance kernel gives
about the data. A constant length-scale may not be used with LIDAR
data due to non-smoothness of collected point cloud. Different
methods are proposed to adjust length scales locally for
non-stationary covariance functions by assuming an exact functional
relationship for length-scale values \cite{Zhang2017, Fuglstad2013,
Fuglstad2015}. The ground segmentation method proposed by
\cite{Chen2014} assumes the length-scales to be a defined function
of line features in different segments. This is not
sufficient because no physical background is considered for the
selection of functional relationship and this function might change
and fail to describe the underlying data in different locations.

In this paper, two Gaussian processes are considered to jointly
perform the ground segmentation task, one to model height of the
ground and the other to model length-scale values. A latent Gaussian
process is set on the logarithm of length-scale values. Point
estimates of length-scale values at each particular ground candidate
location is calculated using a multi-task hyper-parameter learning
scheme. Local estimation of length-scale values enables the method
to consider both flat and sloped terrains. Furthermore, a
whole-frame intuition about ground quality of each frame is injected
into the optimization task by special treatment of selection process
for pseudo-input set. Proposed method is tested on KITTI \cite{Geiger2013} data set
and is shown to outperform similar Gaussian process based ground
segmentation methods.

\section{Radial Grid Map}
 A radial grid mapping is performed on the LiDAR's three-dimensional point cloud. The point cloud data is segmented into $M$ different segments. Then each segment is divided into $N$ different bins.
\begin{figure}[ht!]
\begin{center}
 \includegraphics[scale =0.2,page=2]{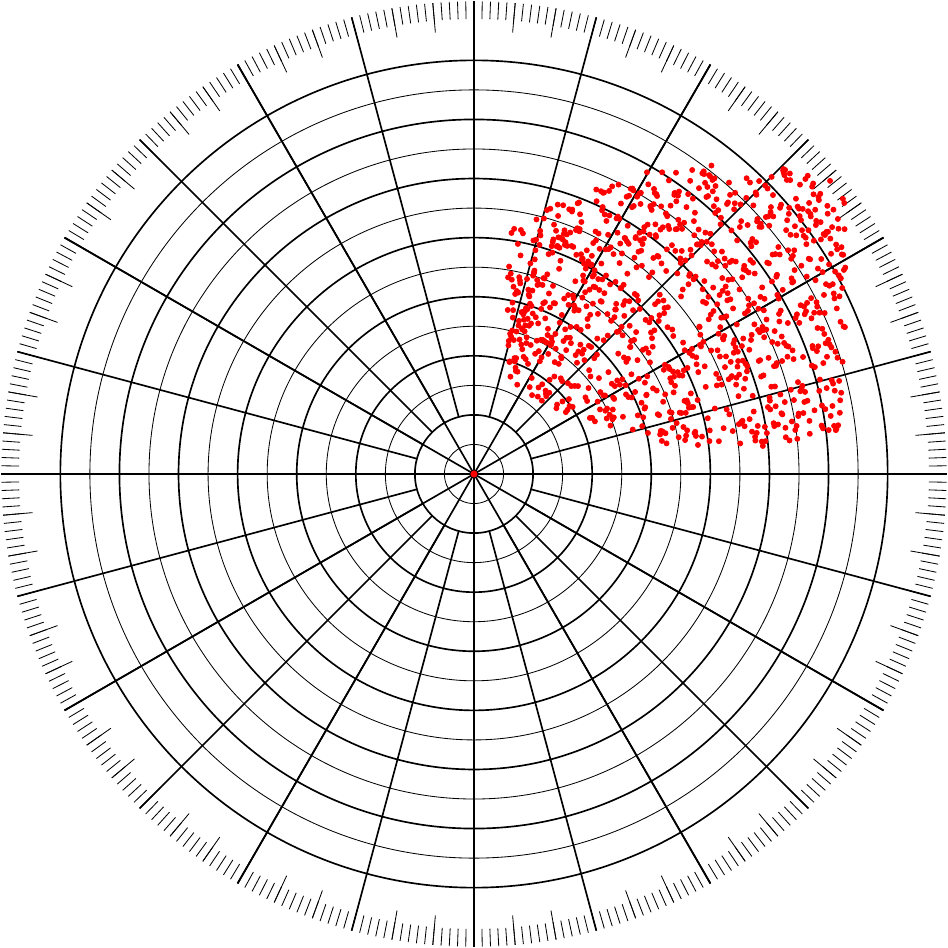}
 \label{fig:gridmap}
  \caption{A Two-dimensional Schematic of The Polar Grid Map}
  \end{center}
 \end{figure}
 The set of all the points in the $n_{th}$ bin of the $m_{th}$ segment is depicted by $P_{b_n^m}$ which covers the range $(r_n^{min}, r_n^{max})$. In order to reduce the computation effort the set $P_{b_n^m}^\prime$ is constructed that contains the corresponding two-dimensional points:
\begin{equation}\label{equation:four}
P_{b_n^m}^{\prime} = \{ p_i^{\prime} = (r_i,z_i)^T \; |\; p_i \in
P_{b_n^m}\},
\end{equation}
where, $r_i = \sqrt{x_i^2+y_i^2}$ is the radial range of corresponding points. The \textit{ground candidate set} $PG_m$, being the first intuitive guess for
obtaining initial ground model is constructed by collecting the
point with the lowest height at each bin as the ground candidate in
that bin \cite{Chen2014}:
\begin{equation}
PG_m = \{p_i^\prime \; | \; p_i^\prime \in P_{b_n^m}^{\prime}, z_i =
\min(\mathcal{Z}_n^m)\},
\end{equation}
where, $\mathcal{Z}_n^m$ is the set of height values in $P_{b_n^m}^\prime$. Furthermore, in each bin a vertical segmentation is applied. Each bin is divided into $j$
vertical segments spread from minimum height $z_{min}^{m,n}$ to
$z_{max}^{m,n}$. Then the number of points in each of these vertical
segments are calculated and averaged on the range to obtain
$\tilde{\rho}_n^m$.
\begin{equation}\label{equation:fourPrime}
\tilde{\rho}_n^m = {\sum_{i=1}^{j}(\rho_i)}/j 
\end{equation}
Where $\rho_i$ is obtained by dividing the number of points in each vertical segment by the area of that segment.

\section{Ground Segmentation by Gaussian Process Regression with Local Length-scale Estimate}
Gaussian processes ($\mathcal{GP}$) are stochastic processes with any finite number of their random variables being jointly Gaussian distributed. In this paper Gaussian process regression is utilized as a tractable method to put prior distributions over nonlinear function that relates ground model to the radial location of points.

Although $\mathcal{GP}$s are very powerful methods to perform Bayesian inference, they fail to consider \textit{non-stationarity} and \textit{local-smoothness} of the data in their general form. In \cite{Mehrabi2021} the non-stationarity of the LiDAR point cloud for the ground segmentation task is addressed by using a non-stationary covariance function as the $\mathcal{GP}$s kernel. The length-scales of the proposed kernel is further obtained by a physically-motivated line extraction algorithm which enables the method to perform well both for the flat and sloped terrains.
\begin{figure}[ht!]
\begin{center}
 \includegraphics[scale =0.17]{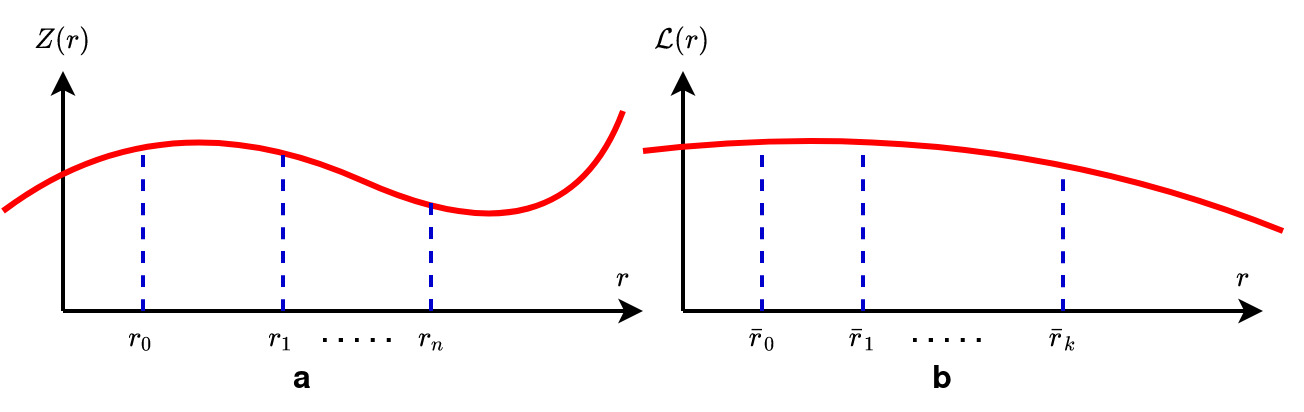}
  \caption{A Schematic of the Two Gaussian Processes. a) The Gaussian process for the ground height estimation b) The Gaussian process for the length-scale estimation.}
   \label{fig:twogp}
  \end{center}
 \end{figure}
The LiDAR data also inherits \textit{input-dependent smoothness} meaning that its data does not
bear smooth variation at every part and direction of the environment, thus the stationarity assumption fails to fully describe ground segmentation task. Therefore, covariance functions with constant
\textit{length-scales} are not suitable for LIDAR point cloud since flat grounds must have a larger length-scale than a rough ground.
\subsection{Problem Definition}
Nonlinear Gaussian process regression problem is to recover a functional dependency of the form $y_i=f(x_i)+\epsilon_i$ from $n$ observed data points of the training set $\mathcal{D}$. In this paper two different functional dependency is to recover: The ground height $h_i=z(r_i)+{\epsilon_z}_i$ and the length-scale values $l_i=\mathcal{L}(r_i)+{\epsilon_l}_i$. The set
$PG_m \{ (r_i,z_i)\}$ contains all of two-dimensional ground candidate points in the segment $m$ that the ground model will be inferred based on. 

The aim of the algorithm is to obtain the predictive ground model $P(z_*|\mathcal{R}_*,\mathcal{R}_m, \theta)$: 
\begin{align}
P(z_*|\mathcal{R}_*,\mathcal{R}, \theta) = \iint P\big(z_* | \mathcal{R}_*,\mathcal{R}, exp(\mathcal{L}_*), exp(\mathcal{L}), \theta_z\big) \nonumber \\
P\big(\mathcal{L}_*, \mathcal{L} | \mathcal{R}_*,\mathcal{R}, \bar{\mathcal{L}},\bar{\mathcal{R}}, \theta_l\big)\,dl\,dl_*
\label{eq:intractablepredictive}
\end{align}
Where $z_*$ represents the ground height prediction at the arbitrary location $r_* \in \mathcal{R}_*$.The parameter $\mathcal{R}$ represents the training data set. $\mathcal{L}_*$ is the mean prediction of the length-scale value at input location $\mathcal{R_*}$ and $\mathcal{L}$ is the mean prediction of length-scale at the training data set's locations. $\theta_z$ and $\theta_l$ represents the hyper-parameters. 

As the marginalization of the predictive distribution of Eq(\ref{eq:intractablepredictive}) is intractable and \textit{Monte Carlo} methods are not efficient for the application, the most probable length-scale estimate is obtained: 
\begin{align}
P(z_*|\mathcal{R}_*,\mathcal{R}, \theta) = P\big(z_* | \mathcal{R}_*,\mathcal{R}, exp(\mathcal{L}_*), exp(\mathcal{L}), \theta_z\big) 
\label{eq:predictive}
\end{align}
Since the length-scales are independent latent variables in the combined regression model, making predictions amounts to making two standard $\mathcal{GP}$ predictions. Thus, two separate $\mathcal{GP}$s are assumed to model the ground segmentation task: $\mathcal{GP}_z$ and $\mathcal{GP}_l$. The $\mathcal{GP}_z$ is assumed to model the functional relation of the ground heights with the radial distance of the points with $(\mathcal{L}_*, \mathcal{L})$ treated as fixed parameters and $\mathcal{GP}_l$ is assumed to model the length-scales of the $\mathcal{GP}_z$s kernel function. A schematic of these two processes are depicted in the Figure \ref{fig:twogp}. 
\subsection{Problem Formulation}
\paragraph{The Gaussian Process for The Ground heights} The $\mathcal{GP}_z$ is defined as follows:
\begin{equation}
\mathcal{GP}_z: z(r) \sim \mathcal{GP}\big( \bar{z}(r),k(r,r^\prime)\big)
\end{equation}

With $z(r)$ being the height values at the location $r$, $\bar{z}(r)$ being the mean value of the height at location $r$ and $k(r,r^\prime)$ being the covariance kernel. Predictive distribution of measurement process can be addressed
after obtaining local point estimate of length-scales on locations
of ground candidate points. Predictive distribution
$P\big(z_*|\mathcal{R}_*,\mathcal{R},exp(\mathcal{L}_*), exp(\mathcal{L}), \theta_z\big)$ enables the prediction of $z^*$ value
at arbitrary locations $r^*$ at  each point cloud frame:
\begin{equation}
\mu_{\mathcal{GP}_z} = \bar{z} = K(r^*,r)^T\bigg[K(r,r)+
\sigma_n^2I\bigg]^{-1}z,
\end{equation}
\begin{equation}
\textrm{cov}_{z_*} = K(r_*,r_*) - K(r_*,r)\bigg[K(r,r)+
\sigma_n^2I\bigg]^{-1}K(r_*,r)^T.
\end{equation}

in which, K is the covariance matrix for $\mathcal{GP}_z$ and
$\sigma_n^2$ is the measurement noise. The non-stationary covariance
kernel from \cite{Paciorek2003} is chosen to represent correlation
of measured data points:
\begin{eqnarray}\label{equation:nonscov}
&K(r_i,r_j) = \sigma_f^2.\big[\mathcal{L}_i^2\big]^{\frac{1}{4}}
\big[\mathcal{L}_j^2\big]^{\frac{1}{4}}
\big[\frac{\mathcal{L}_i^2 + \mathcal{L}_j^2 }{2} \big]^\frac{-1}{2}\nonumber \\
& \times  \exp\bigg(\frac{-(r_i-r_j)^2}{[  \mathcal{L}_i^2 +
\mathcal{L}_j^2]}\bigg),
\end{eqnarray}
where, $\mathcal{L} \in \mathbb{R}$ is length scale for every data
point.
The predictive distribution of the height
$z_*$ at the arbitrary test input location $r_*$ are obtained using
predictive ground model.
\paragraph{Local Length-Scales Estimation}
Mean value prediction on latent Gaussian process $\mathcal{GP}_l$ results point estimates of length-scale values $\mathcal{L}$. The predictive distribution $P(\mathcal{L}^*|\mathcal{R}^*,\mathcal{R})$
enables point-wise estimation of Length-scale values. regarding to each data point, are calculated by
mean prediction of latent process:
\begin{equation}
\mathcal{GP}_l :\Rightarrow P(\mathcal{L}_*|\mathcal{R}_*,\mathcal{R})
\sim \mathcal{N}(\mathcal{L}_*,\sigma_*^2),
\end{equation}
\begin{equation*}
\mathcal{L}_*=\mu_{\mathcal{GP}_l}=\log \mathcal{L}=(\bar{K}(r,\bar{r}))^T[\bar{K}(\bar{r},\bar{r})+\bar{\sigma}_n^2I]^{-1}
\bar{\mathcal{L}},
\end{equation*}
\begin{equation}
\Rightarrow\mathcal{L} = \exp [(\bar{K}(r,\bar{r}))^T
[\bar{K}(\bar{r},\bar{r})+\bar{\sigma}_n^2I]^{-1}\bar{\mathcal{L}}],
\end{equation}
in which, $\mathcal{L}^*=\mu_{\mathcal{GP}_l}$ is the mean value
prediction of the logarithm of the length-scale at the desired locations
$\bar{R}^*$. To obtain length-scale values for each data point, desired locations are adjusted to ground candidate data points at
each segment. Matrix $\bar{K}(\bar{r},\bar{r})$ and vector
$\bar{K}(r,\bar{r}))$ are corresponding co variances. Furthermore,
Covariance parameters for latent process are defined by stationary,
squared exponential covariance kernel:
\begin{equation}
\Rightarrow\bar{k}(\bar{r}_i,\bar{r}_j) =
\bar{\sigma}_f^2\exp\bigg(-\frac{1}{2}\frac{(\bar{r}_i -
\bar{r}_j)^2}{\bar{\sigma}_l^2}\bigg).
\end{equation}
The $(\bar{\mathcal{L}}, \bar{\mathcal{R}})$ is the training set for the length-scale process. These locations of latent kernel are often chosen randomly in the variable space defined by $\mathcal{R}$. In this paper in each segment a \textit{pseudo input selection} algorithm is performed to obtain these latent training data points.

\subsection{Pseudo-Input Selection}
The input locations for the latent Gaussian process are obtained due to the physical qualities of the ground surface in each segment. A common assumption using LIDAR sensors is that measurements coming from a certain surface must be somehow more related to each other. This assumption yields that if certain area of point cloud data resembles a surface, covariance related characteristics must be similar in
that area and differ from other neighboring data points which does
not show dependency to the same surface. 

Therefore if some measure is introduced for all the points from certain hypothetical surface in point cloud data, this measure can represent a fair measure of length-scale value for that certain area of the data, loyal to the
shared surface. \subsection{Physically-Motivated Line Extraction}
Different line extraction algorithms are being utilized in different
robotics applications with some of them being more generally
accepted and utilized. Although these algorithms are widely used,
some applications need to use different versions of
them~\cite{Chen2014}, \cite{Siadat1997}. Line
extraction algorithms are previously used in different ground
segmentation methods to enable the distinction of the ground and
obstacle points. Thus, although these methods are effective in segmenting near-flat ground points, they fail to properly
recognize ground points coming from sloped ground sections or
gradient roads. \cite{Chen2014} states that despite using a
non-stationary covariance function as kernel, their method does not
work in the existence of sloped terrains. This can be due to the usage of the Incremental line
extraction algorithm which is elaborated in \cite{Nguyen2005} and
utilized in \cite{Chen2014} to estimate ground surface. The incremental line extraction algorithm lacks the
efficiency needed for the detection of sloped terrains. Thus a physically
motivated line extraction algorithm along with a two-dimensional line fitting method is introduced which is
intuitively compatible with what happens in real-world urban
scenarios. The proposed line extraction algorithm relies on the fact
that in urban structures, the successive lines of each laser scan
should be considered independently. For example, if some structures
are found in the data that shows a $3^{\circ}$ slope for a range of
radial distance ($2m$) and the algorithm is decided that the last
point of this series is a "critical point", the parameters of the
next line will start to construct from scratch and without
dependency to previous segments. This is because the ground
candidate points are successive points coming from different bins.
Therefore, a sudden change of structure is more important than the
overall behavior of some cluster of points as they may be related to
a starting point of an obstacle or sloped ground.

\paragraph{Definition of Critical Points}
To overcome the problem of sloped terrain detection in the ground
segmentation task, the proposed line extraction algorithm operates
based on finding some \textit{\textbf{critical points}} among the
ground candidate points set in each segment $PG_m$. The critical
points are defined to be the points at which the behavior of the
successive ground candidate points change in a way that can be
flagged as \textit{unusual}. This unusual behavior happens in the
areas that the ground meets the obstacle or as well as the areas at
which the road starts to elevate during a slope or gradient section
of its. These critical points, therefore, partition each segment
into different sections between each two successive critical points.
The 2D points between two successive critical points form a
\textbf{\textit{line-segment}}. These line-segments should be chosen
carefully as they play a significant role in the line extraction
algorithm and the further interpretation of the ground. On the other
hand, while large sloped line-segments relate to non-ground
structures, the low sloped ones may relate to the ground. Therefore,
The conditions listed below must be met for a ground point to be
considered as a critical point:
\begin{itemize}
    \item \textit{The slope of fitted line}: the slope of the fitted line for the potentially ground-related line segments must be greater than a threshold $\zeta_b$.
    \item \textit{Distance from point to the fitted line:} distance
    from points of each line-segment to the fitted line must be less
    than a given threshold $\delta$.
    \item \textit{Horizontal Distance of the points:} The horizontal distance of each two successive points must not exceed a given threshold $\zeta_m$. This is set to prevent including breakpoints.
    This threshold is set concerning each segment and about the radial size of each segment.
    \item \textit{Smoothness of $\tilde{\rho_n^m}$}: The average number of vertical points must not have a sharp change during each line-segment.
\end{itemize}
\paragraph{Distance-oriented Line Regression}
The distance-oriented line regression method is utilized as the core
line fitting algorithm. The standard least square method breaks down
when the slope of the line is almost vertical or when the slope of
the line is large but finite. The least-square method assumes that only the dependent value is subject to error
thus the distance of the points to the estimated value $(\hat{y}-y)^T(\hat{y}-y)$ is utilized. This assumption makes the algorithm
very sensitive to the position of the independent value especially
in larger slopes. In
these cases, a small inaccuracy in the value of the regressor will
lead to greater uncertainty in the value of the regressed parameter
if the least square is used as the line fitting algorithm. On the other hand, in many applications such as ground segmentation,
this assumption fails to be true as both coordinates are subject to
errors. Thus, the least square method is insufficient for sloped
terrain ground point detection as the method tends to obtain critical points in
areas with larger slopes and the accuracy of detection are of high
importance here. The distance-oriented line regression method takes
the exact distance of the points to the line into account. The orthogonal least square algorithm is utilized as
an alternative for the least square method. The method assumes that
both parameters have the same error while this assumption is not
valid for line-segment extraction as the 2D value is derived by
implementing manipulation on original 3D data. Furthermore, the
Non-stationarity assumption of the method denies any similarity of
errors in both directions. The mathematical model $\eta=\alpha+\beta \xi$ is assumed to describe the linear relationship of two
underlying variables. If both of the variables are observed subject
to a random error, the relation of these measurements are as
follows:
\begin{eqnarray}
&r_i = \xi_i + \epsilon_{r_i} \\
&z_i = \eta_i + \epsilon_{z_i} = \alpha + \beta \xi_i + \epsilon_{z_i}
\end{eqnarray}
In which $\epsilon_{r_i}\sim\mathcal{N}(0, \sigma^2)$ and
$\epsilon_{z_i}\sim\mathcal{N}(0, \lambda\sigma^2)$. The maximum
likelihood estimate for $\alpha$ and $\beta$ is derived. The
"log-likelihood" has the following form:
\begin{eqnarray}
&&L = -\frac{n}{2}log(4\pi^2)-\frac{n}{2}log(\lambda\sigma^4)\nonumber \\
&&-\frac{\Sigma_{i=1}^n(r_i-\xi_i)^2}{\sigma^2}-\frac{\Sigma_{i=1}^n(z_i-\alpha
-\beta\xi_i)^2}{2\lambda\sigma^2}
\end{eqnarray}
Differentiating the log-likelihood function with respect to
${\alpha, \beta, \xi, \sigma}$ and solving for $\frac{\partial
L}{\partial \star} = 0$ yields the numerical estimate of the line parameters.
   \begin{center}
   \begin{figure*}[ht]
      \centering
      \framebox{\parbox{0.6\textwidth}{
  \begin{minipage}[b]{0.5\linewidth}
    \includegraphics[width=0.9\textwidth]{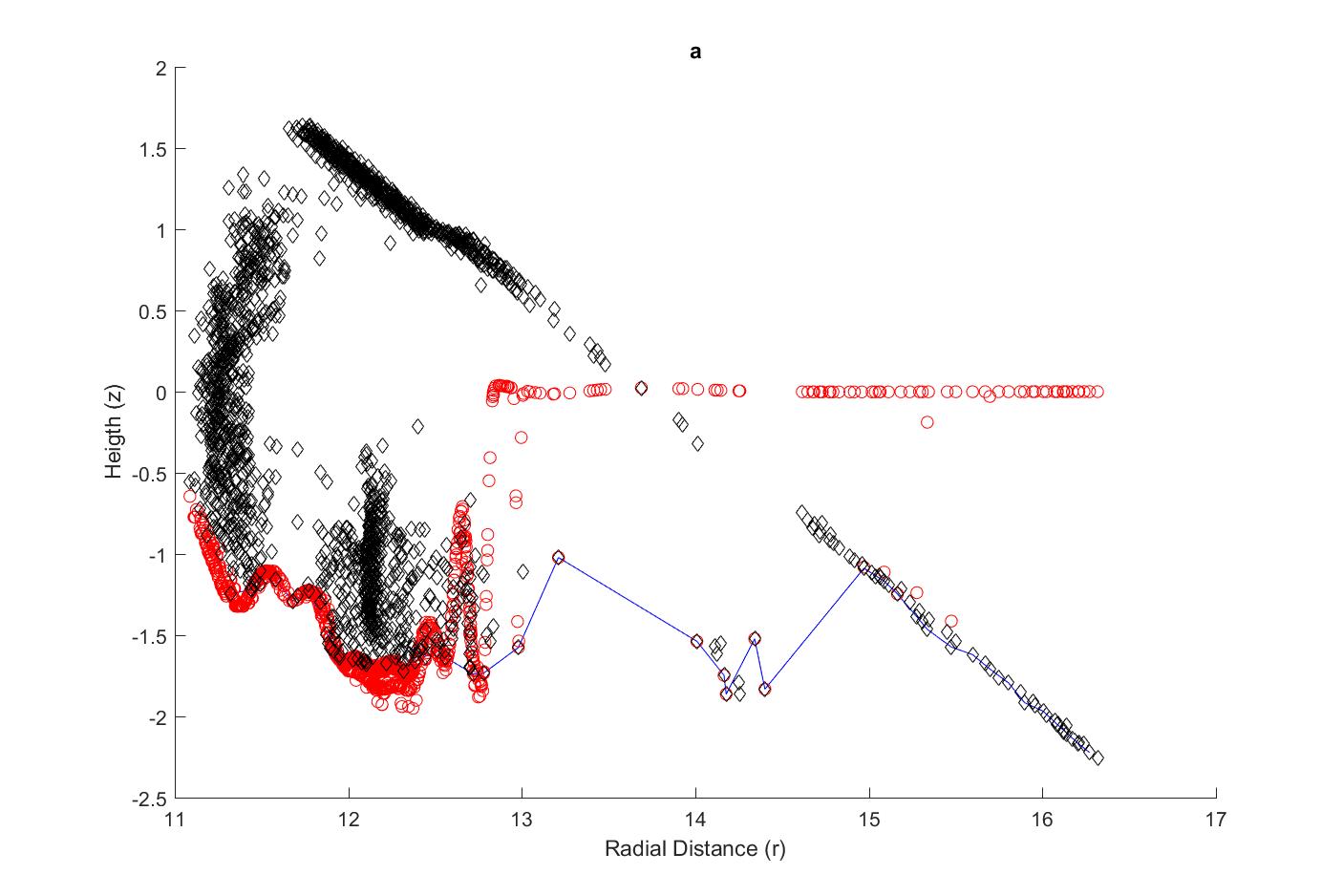}
          \vspace{0.5cm}
  \end{minipage}
  \hfill
  \begin{minipage}[b]{0.5\linewidth}
    \includegraphics[width=0.9\textwidth]{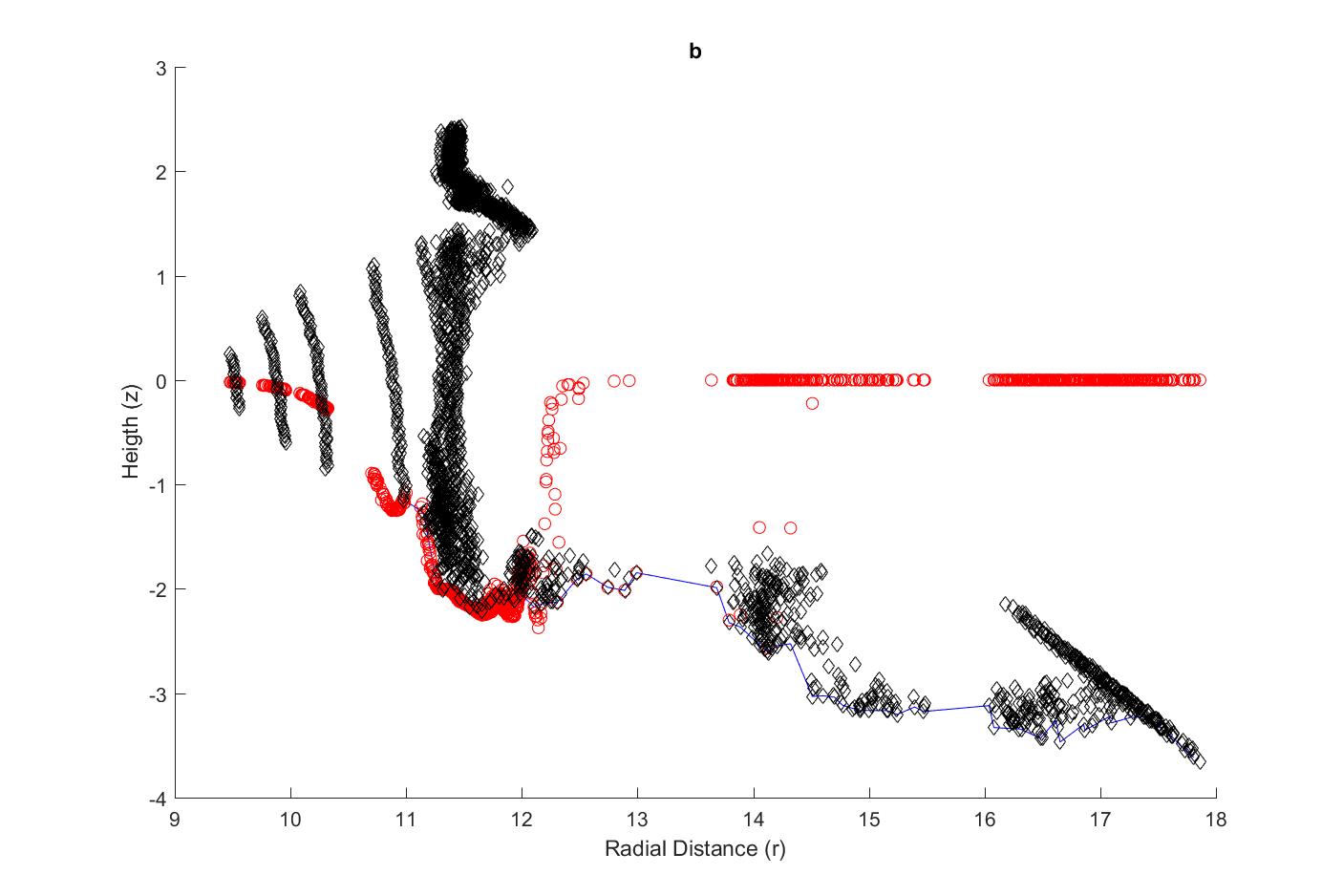}
      \vspace{0.5cm}
  \end{minipage}
   \begin{minipage}[b]{0.5\linewidth}
       \includegraphics[width=0.9\textwidth]{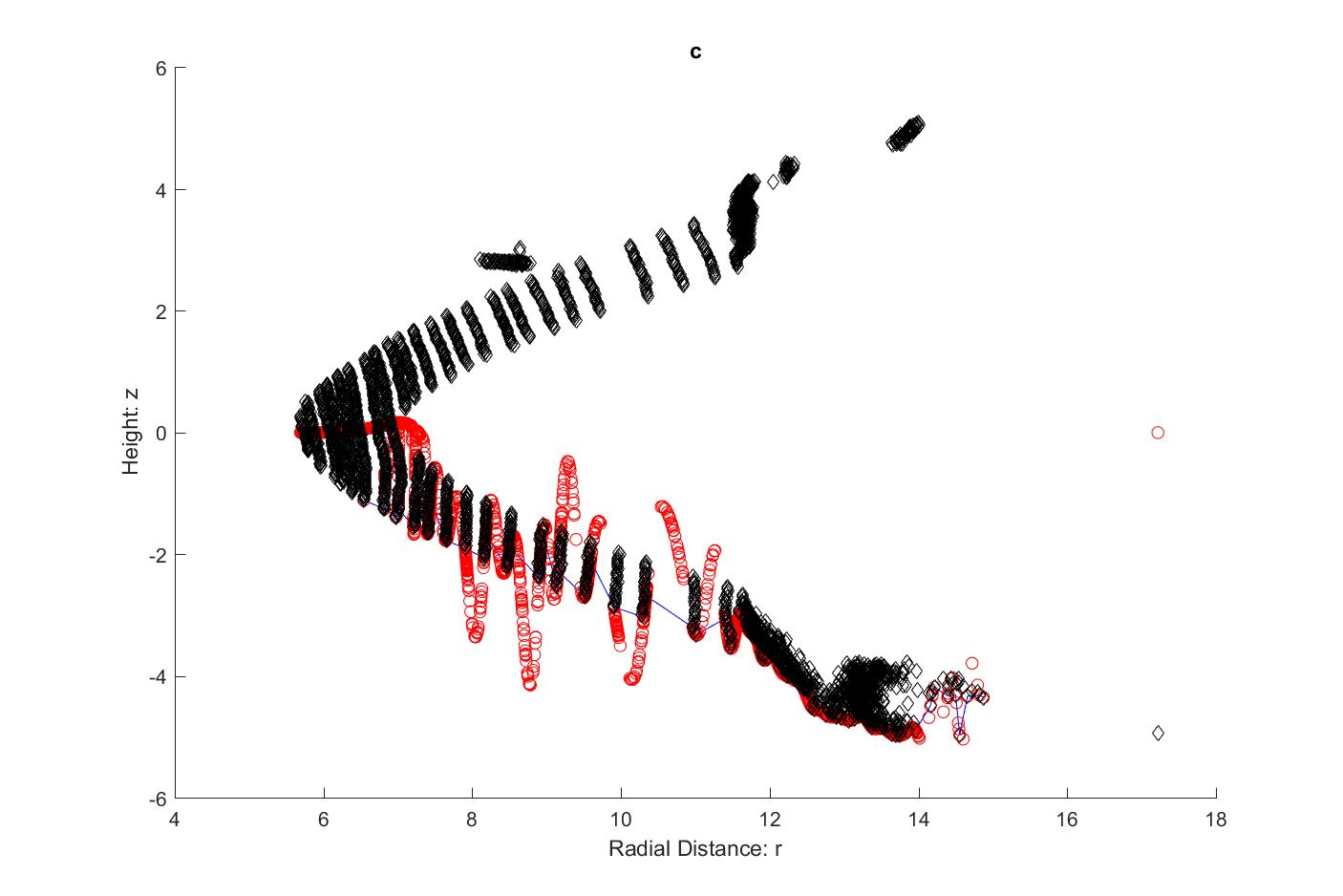}

  \end{minipage}
     \begin{minipage}[b]{0.5\linewidth}
    \includegraphics[width=0.9\textwidth]{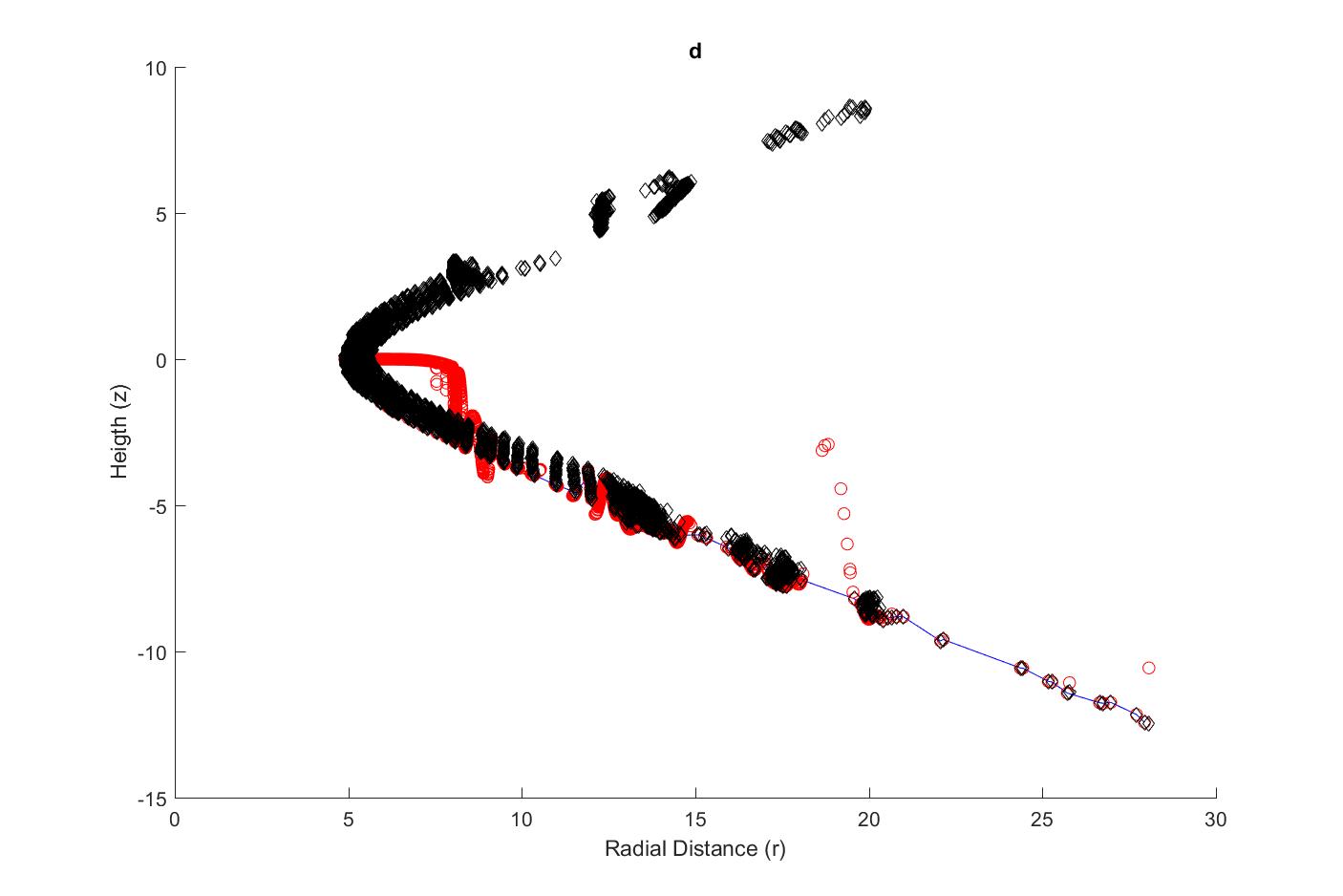}
  \end{minipage}
     \caption{Ground segmentation results for two different adjacent segments:
      Red circles represent estimated ground for data points and black squares are the raw data. Blue line is the ground candidate set for each segment.}
           \label{fig:one}
}}
   \end{figure*}
   \end{center}
\subsection{Learning Hyper--Parameters}
The behavior of the proposed model and its ability to adjust itself
to physical realities of the environment is directly effected by the
values of hyper--parameters. Often in real world applications there
is no exact prior knowledge about hyper--parameters value and they
must be obtained from data. Hyper--parameters $\theta =
\{\sigma_f,\bar{\mathcal{L}}, \sigma_n, \bar{\sigma}_f,
\bar{\sigma}_l, \bar{\sigma}_n\}$ are mutually independent variables
to allow the gradient-based optimization to hold its credibility.

\emph{Log Marginal Likelihood:} Marginal likelihood or evidence
$P(y|\mathcal{X})$ is the integral of likelihood times the prior. Logarithm of marginal likelihood is obtained under Gaussian process
assumption that the prior is Gaussian $f|\mathcal{X}\sim
\mathcal{N}(0,K)$:
\begin{equation}
\log P(f|\mathcal{X}) =
-\frac{1}{2}f^TK^{-1}f-\frac{1}{2}\log|K|-\frac{n}{2}\log(2\pi)
\end{equation}

The \textit{maximum a posteriori} argument is used to find the
hyper--parameters for ground segmentation. The hyper--parameters
that maximize the probability of the likelihood of observing $z$
given $R$, are assumed to be valid values for our regression:
\begin{eqnarray}
&\log P(\mathcal{L}|z,R,\theta)=  \nonumber \\
&\log P(z|R,\exp(\mathcal{L}),\theta_z)+ \log
P(\mathcal{L}|R,\bar{\mathcal{L}},\bar{R},\theta_l)
\end{eqnarray}
Gradient--based optimization methods are used in order to find the
corresponding solutions.

\paragraph{Whole-Frame Objective Function} If optimization process is to
be effective enough, it shall take all the segments into account to
form an \textit{objective function}. This would be an example of
\textit{multi-task} regression problem. The objective function in
multi-task problems is equal to the sum of all objective functions
regarding to different tasks of the problem that share the same
hyper-parameters. Therefore, we assume that all the segments share
the same \textit{hyper--parameters} and establish a global view to
our frame data in order to have the results to be whole-frame
inclusive.
\begin{eqnarray}
L(\theta) & = & \sum\limits_{m=1}^M \bigg(\log P(\mathcal{L}^m|z^m,R^m,\theta)\bigg)  \\
 &  = &  -\frac{1}{2}\bigg[\sum\limits_{m=1}^M\bigg((z^m)^TA_m^{-1}z^m\bigg)+\log\bigg(\prod\limits_{m=1}^M |A_m|\bigg) \nonumber \\
 & + &\log\bigg(\prod\limits_{m=1}^M |B_m|\bigg)+\log(2\pi)\bigg(\sum\limits_{m=1}^M(n_m+\bar{n}_m)\bigg)\bigg]\nonumber
\end{eqnarray}
where $A_m:=K_m(r,r)+\sigma_n^2I$ and
$B_m:=\bar{K}_m(\bar{r},\bar{r})+\bar{\sigma}_n^2I$ are the
corresponding covariance functions of Gaussian processes in each
segment.  Therefore, in every segment the ground candidate set with
assigned covariance kernels form the segment-wise objective
function. Sum of all segment-wised objective functions will form the
whole-frame objective function. For the whole-frame objective
function to hold its credibility, the hyper-parameters of the
regression task must be shared among all segments. Therefore,
interpretation of \textit{length-scale} parameters must be redefined
in pseudo-input selection, for  all segments to be able to share the
hyper-parameter $\bar{\mathcal{L}}$.

\begin{center}
\begin{algorithm}[t]
  \scriptsize
  \caption{Line Extraction in segment m}
\begin{algorithmic}[1]
\State \textbf{INPUT:} $M$, $N$, $\zeta_m$, $\zeta_b$
\State \textbf{OUTPUT:} $PG_m$, $L_m$
\State $k=0$
 \For { $j = 1 \; to \; M $ }
    \For { $i = 1\; to \; Size(PG_m(j))$ }
        \If {
        $P_{b_n^m}^\prime \neq \emptyset $}

              \State $d_L = size(PG_m(j))$
              \If {$|d_L| \geq 2 $}
                \State $s_i = 2$
                \For {$n = si \; to \; d_L$}
              \State $(a_k,b_k,c_k)=Fitline(d_L)$
\If{
              \begin{tiny}
$|\frac{a_k}{b_k} - \frac{a_{k-1}}{b_{k-1}}|\leq \zeta_m$ \textbf{or}
$| \frac{c_k}{b_k} - \frac{c_{k-1}}{b_{k-1}} | \geq \zeta_b$
              \end{tiny}
              } \EndIf

              \State {$s_i = n+2$}
              \State { \textbf{goto} Line: 10 }
              \State {\textbf{else}}
                          \State $L_m(j) = L_m(j) \cup \{a_k,b_k,c_k\}$
              \EndFor
               \EndIf
               \State { $k = k+1$ }
            \State $d_L = \emptyset$
            \EndIf
            \EndFor
            \EndFor
\end{algorithmic}
\label{algo:line}
\end{algorithm}
\end{center}

\subsection{Gradient Evaluation}
The gradients of log marginal likelihood objective function
$L(\theta)$ are calculated analytically using,
\begin{eqnarray}
\frac{\partial L(\theta)}{\partial\ast} =
\frac{1}{2}\sum\limits_{m=1}^M\bigg((z^m)^TA_m^{-1}
\frac{\partial A_m}{\partial \ast}A_m^{-1}z^m\bigg)- \nonumber \\
\frac{1}{2}\sum\limits_{m=1}^M\textbf{tr}
\bigg(A_m^{-1}\frac{\partial A_m}{\partial \ast}\bigg)
-\frac{1}{2}\sum\limits_{m=1}^M\textbf{tr}\bigg(B_m^{-1}
\frac{\partial B_m}{\partial \ast}\bigg)
\end{eqnarray}
It is then obvious that if we calculate the $\frac{\partial
A_m}{\partial \ast}$ and $ \frac{\partial B_b}{\partial \ast}$ for
all divided segments, the remaining calculations are found straight
forward.
\subsection{Ground Segmentation Algorithm}
The final ground segmentation algorithm is represented in algorithm \ref{algo:ground}. In every segment \textit{m}, a candidate two-dimensional ground
point set $PG_m$ is constructed for every LIDAR frame which can be
contaminated by obstacle points as outliers. The typical Gaussian
process regression task assumes that all of the data in $PG_m$ is
ground points with a few outliers. Ground segmentation problem is
formulated as obtainment of one regression model with the ability of
outlier rejection for each segment in radial grid map, also an
iterative learning method is adapted to build the local ground model
in every segment which benefits at the same time from both desirable
approximation ability and outlier rejection. 
\begin{algorithm}
 \scriptsize
\caption{Ground Segmentation}
\textbf{Input}: $P_t = \{p_1,p_2,...,p_l\},M,N,B,\bar{\sigma}_f,\bar{\sigma}_n,\bar{\sigma}_l, \bar{\mathcal{L}}$, \\ $\sigma_f,\sigma_n,\zeta_g$ \\
\textbf{Output}: Label of each point $p_i \; i=1,2,...,l$
\begin{algorithmic}[1]
\State $(PG_m,P_{b_n^m}^\prime) = RadialGridMap(P_t,M,N)$
\For{i = 0 : M-1}
\For{j = 0 : \textit{Size}($P_{b_n^i}^\prime$)}
\State  $L_j$ = \textit{LineExtraction}($P_{b_n^i}^\prime$)
  \If {$Size(PG_i) > 0$ }
\State  ($\bar{y}_i$, $\bar{r}_i$) = \textit{PseudoInputSelection}($PG_i$)
\State  $\mathcal{L}$ = \textit{\textbf{$GP_l$}}($\bar{r}_i, \bar{y}_i, \bar{\sigma}_f,\bar{\sigma}_n,\bar{\sigma}_l, \bar{\mathcal{L}}$)
\State  $\mu_*^{l}\leftarrow \mu_*^{ij}$ = \textit{\textbf{$GP_z$}}($r_*,\mathcal{L},\sigma_f,\sigma_n$)
\EndIf
\EndFor
\EndFor
\For{i = 0 : l }
\If {$|z_i - \mu_*^{i}| > \zeta_g $} label($p_i$) = Obstacle
\Else \; label($p_i$) = Ground
\EndIf
\EndFor
\end{algorithmic}
\label{algo:ground}
\end{algorithm}
This algorithm actually starts with receiving a 3D scan of
environment as a set of point clouds $P=\{P_1,...,P_k\}$ which is
consisted of $P_t$ frames at time $t$ and outputs the label of each
point in each point cloud frame as \textbf{ground} or
\textbf{obstacle}. The parameter $\zeta_g$ represents predefined value for ground loyalty. Points with this height distance from continuous estimation of the ground are considered as obstacle points. 
\section{Implementation Results}
For implementation purposes the input space is divided into $M=180$
different segments covering a 2--degree portion of the environment
and each segment is divided into $N=120$ bins. For line extraction
purpose $\zeta_m$ is set to $0.02$ radians and $\zeta_b$ is set to
$15^{cm}$, therefore, any line structure with height difference less
than $0.15^m$ and less than $0.02$ radians deviation from mean angle
are assumed to be loyal to a unique line. Regression prediction
error is validated using standardized mean squared error by
\begin{equation}
\textrm{SMSE} \equiv \; \frac{1}{n}
\sum\limits_{i=1}^n\big[\textrm{var}(y)^{-1}(y_i-\mu_i)^2\big]
\end{equation}
All the results are obtained on a laptop with Intel Core i7 6700HQ
processor and simulations are implemented using point cloud library (PCL) and with C++, while figures are prepared by MATLAB.

Figure \ref{fig:one} shows two different adjacent segments and their
ground segmentation result, in which, red circles represent the
estimated ground for data points, while black squares are the raw
data. As it can be seen in the first row, Figure \ref{fig:one}~$a$
and $b$, predicted ground for two adjacent segments have a flat
structure, although segment $a$ shows a sloped structure in that
area and segment $b$ shows an uneven depth. While the ground
candidate set showed in the figure by connected blue line proposes
other structure, ground segmentation algorithm predicts a flat
ground for this area as a result of considering all the segments
together and having a whole-frame. This results shows that general
point of view does not obey a segment--wise logic. On the other hand
in segments shown in part $c$ and $d$ of this figure, although data
in segment $d$ covers a wider region, algorithm tends to estimate
detailed structure of ground in both segments regardless of what
local ground candidate set may impose based on segment--wise logic.
This feature enables the algorithm to truly recognize ground in
radial distances ranged from $6$ to $8$, although segment $d$ offers
a more smooth ground shape from its data in that range.
   \begin{table*}[ht]
    \scriptsize

\begin{center}
\caption {Ground Prediction Results}
\begin{tabular}{cccccccccc}
     \toprule \toprule
    {$n_{PG_m}/n_{\mathcal{L}}$} & {$\sigma_f^2$} & {$\sigma_n^2$} &
    {$\bar{\sigma}_f^2$} & {$\bar{\sigma}_l^2$} & {$\bar{\sigma}_n^2$}
    & {$\textrm{SMSE}_{Z-\mathcal{L}}$} & {$\textrm{SMSE}_{Z}$} & {$t_{Z-\mathcal{L}}(s)$} & {$t_Z(s)$}\\
         \midrule
    36/23 & 0.3663 & 0.8607 & 1.415e+6 & 2.250e+6 &
    1.187e+6 &  2.4258e-27 & 1.4978e-23 & 0.5432 & 0.2335 \\
  \midrule
    52/30 & 0.3117 & 0.7207 & 4.729e+6 & 1.347e+6 &
    2.058e+6 &  1.2466e-28 & 1.1231e-23 & 1.0045 & 0.4356 \\
  \midrule
      68/16 & 0.4896 & 1.1544 & 1.381e+6 & 3.881e+6 &
      4.501e+6 & 2.7561e-26 & 1.4102e-21 & 0.3283 & 0.1963\\
  \midrule
      81/14 & 0.1103 & 1.6487 & 4.588e+6 & 3.521e+6 &
      4.402e+6 & 4.2903e-19 & 8.6733e-22 & 0.2732 & 0.3412\\
  \toprule \toprule
  \label{table:one}
\end{tabular}
\end{center}
  \end{table*}
Furthermore, ground segmentation method of \cite{Chen2014} is
implemented on the same data. Detailed results of the implementation
is given in Table~\ref{table:one}. The first two rows are corresponding estimated values for figure \ref{fig:one}~$a$
and $b$ with $n_{PG_m}/n_{\mathcal{L}}$ being the number of ground candidate points versus number of selected pseudo-input set. It is important to notice that reducing $n_{\mathcal{L}}$ by weakening pseudo-input selection criteria will increase speed of the given algorithm with the expense of reducing the precision. This enables a trade-off between precision and speed of the algorithm. Given results shows that in these two segments, choosing a large pseudo-input set gives a more precise ground estimation result than the other method while making it slower. However it is still real-time applicable and implementable for urban driving scenes. The last two rows are corresponding results for figure \ref{fig:one}~$c$ and $d$ while it can be seen that by choosing a small pseudo-input set relating to ground candidate set $14/81$ makes our prediction faster while reduces the precision. This comes from the fact that as the size of pseudo-input set gets larger, the gradient-based optimization step becomes more time consuming. The SMSE. hyper-parameters and time values reported in table~\ref{table:one} are recorded for ground segmentation in the corresponding frame while in figure~\ref{fig:one} just selected segments are depicted.

As it is seen in these comparison results the proposed method
denoted by $Z-\mathcal{L}$ outperforms the conventional method in precision,
while being fast enough for urban driving scenarios and robust to the locally changing characteristics
of input point cloud. Calculated SMSE error shows that the proposed
method is more precise than
that of the method given in \cite{Chen2014}, and its speed is
related to the ratio of number of points in pseudo--input set to the
number of points in ground candidate set. The SMSE error for each
segment is reported as the mean value of 100 iterations of ground
prediction. Furthermore, values of hyper-parameters are reported. In
order to consider different scales of covariance kernels, while
scaled gradient-based optimization method is used to find optimal
hyper--parameters.

This real-time applicable ground segmentation procedure may be
applied efficiently in applications, where fast and precise clustering and
ground segmentation is needed to enable further real-time processes
like path planning or dynamic object recognition and tracking especially in rough scenes. Furthermore, taking location-dependent characteristics of non-linear
regression into accounts, enables the method to show better
performance in rough scenes. In addition, the initial guess of
length-scale parameters which is based on fast line extraction
algorithm, increases the time efficiency of the proposed method by
providing physical motivated initial guess for the
$\bar{\mathcal{L}}$ vector. Furthermore, the physical motivated
procedure of choosing each segment's length-scale vector, gives the
method a sense of intuition which is related to the ground quality.
This intuition which is behind choosing the most important
parameters of the regression method, ensures the fair calculation of
ground at each individual frame with respect to its specifications.
\section{Conclusions} A physically-motivated ground segmentation method
is proposed based on Gaussian process regression methodology. Non-smoothness 
of LIDAR data is introduced into the regression task by choosing a non-stationary covariance function represented in equation~\ref{equation:nonscov} for the main process. Furthermore, local characteristics of the data is introduced into the  method by considering non-constant length-scales for these covariance function. A latent Gaussian process is put on the logarithm of the length-scales to result a physically-motivated estimation of local characteristics of the main process. A pseudo-input set is introduced for the latent process that is selected with a whole-frame view of the data and based on the ground quality of the data in each segment by assuming the related correlation of data points that are gathered from same surface.

 It is verified in this paper that the
proposed method outperforms conventional methods while  being realistic, precise
and real-time applicable. Furthermore, presented results shows that
proposed method is capable of effective estimation of the ground in
rough scenes. While the ground structure in the given example in
figure \ref{fig:one} may be assumed to be rough as it contains bumpy
structures and sloped obstacles, the proposed method is capable to
detect ground model efficiently and precisely.

\addtolength{\textheight}{-8cm}   








\bibliographystyle{IEEEtran}
\bibliography{IEEEabrv,gpz.bib}

\end{document}